\RequirePackage{fix-cm}
\documentclass[smallextended]{svjour3}       

\usepackage[square,comma,numbers,sort&compress,sectionbib]{natbib}
\usepackage{paralist}
\usepackage{todonotes}
\usepackage{pgfplotstable} 
\usepackage{lineno,hyperref}
\usepackage[skip=0pt]{caption}
\usepackage{makecell}
\usepackage{multirow}
\usepackage{enumitem}
\usepackage{amsmath}
\usepackage{arydshln}
\usepackage{booktabs}
\usepackage{tabularx}

\hypersetup{urlcolor=blue, colorlinks=true} 
\usepackage{graphicx}
\usepackage{url}
\usepackage[utf8]{inputenc}
\usepackage{wrapfig,lipsum}
\usepackage{graphicx}
\usepackage{pgfplots}
\usetikzlibrary{patterns}
\usepackage{subcaption}
\usepackage{textcomp}
\usepackage{tikz}
\usepackage{pgfplotstable}
\usetikzlibrary{arrows,positioning}
\usepackage{hyperref}
\hypersetup{
 colorlinks,
 citecolor=blue,
 filecolor=black,
 linkcolor=blue,
 urlcolor=black
}

\usetikzlibrary{plotmarks}
\pgfplotsset{compat=1.8}

\smartqed  
\usepackage{graphicx}
\begin{document}

\title{Open Domain Suggestion Mining
\thanks{%
This research was funded by Science Foundation Ireland under grant no. SFI/12/RC/2289 (Insight Centre for Data Analytics), and European Union funded project MixedEmotions (H2020-644632).
Ahold Delhaize,
Amsterdam Data Science,
the Bloomberg Research Grant program,
the China Scholarship Council,
the Criteo Faculty Research Award program,
Elsevier,
the European Community's Seventh Framework Programme (FP7/2007-2013) under
grant agreement nr 312827 (VOX-Pol),
the Google Faculty Research Awards program,
the Microsoft Research Ph.D.\ program,
the Netherlands Institute for Sound and Vision,
the Netherlands Organisation for Scientific Research (NWO)
under pro\-ject nrs.\ 
CI-14-25, 
652.\-002.\-001, 
612.\-001.\-551, 
652.\-001.\-003, 
and
Yandex.
All content represents the opinion of the authors, which is not necessarily shared or endorsed by their respective employers and/or sponsors.
}
}
%

\subtitle{Problem Definition and Datasets}


\author{
Sapna Negi
\and
Maarten de Rijke
\and        
Paul Buitelaar
}

\institute{%
Sapna Negi$^*$ \at
Genesys Telecommunications Laboratories, Inc., Daly City, CA, United States\\
\email{sapna.negi@genesys.com}
\and
Maarten de Rijke \at
Informatics Institute, University of Amsterdam, Amsterdam, The Netherlands\\
\email{derijke@uva.nl}  
\and
Paul Buitelaar \at
Insight Centre for Data Analytics, NUI Galway, Galway, Ireland\\
\email{paul.buitelaar@insight-centre.org}
\and 
$^*$ Work done while at the Insight Centre for Data Analytics, and the University of Amsterdam.
}
\date{Received: date / Accepted: date}

\maketitle

\begin{abstract}
We propose a formal definition for the task of suggestion mining in the context of a wide range of open domain applications. 
Human perception of the term \emph{suggestion} is subjective and this effects the preparation of hand labeled datasets for the task of suggestion mining. 
Existing work either lacks a formal problem definition and annotation procedure, or provides domain and application specific definitions. 
Moreover, many previously used manually labeled datasets remain proprietary. 
We first present an annotation study, and based on our observations propose a formal task definition and annotation procedure for creating benchmark datasets for suggestion mining. 
With this study, we also provide publicly available labeled datasets for suggestion mining in multiple domains.
\end{abstract}


\section{Introduction}
\label{Intro} 

Suggestion mining can be defined as the extraction of sentences that contain suggestions from unstructured text. 
Collecting suggestions is an integral step of any decision making process. 
A suggestion mining system could extract exact suggestion sentences from a retrieved document, which would enable the user to collect suggestions from a much larger number of pages than they could manually read over a short span of time.

\begin{figure}[t]
\centering
\fbox{\includegraphics[scale=0.5]{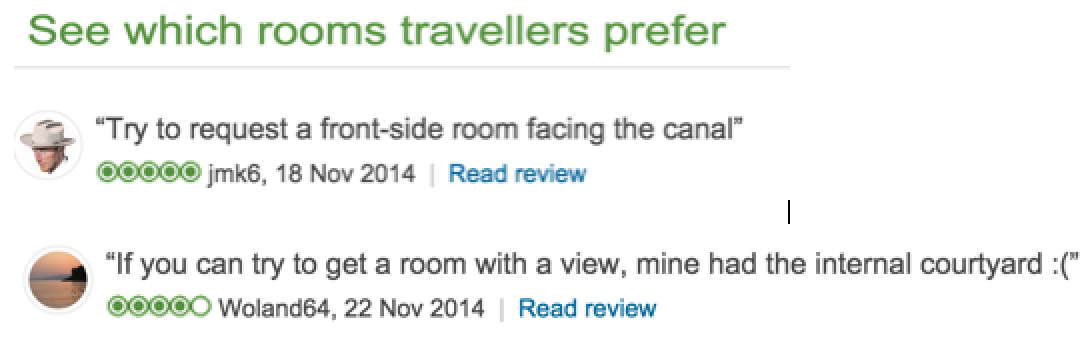}} 
\caption{Manually provided room tips on TripAdvisor}
\label{roomtips}
\end{figure}

Apart from suggestions that relate to general topics, industrial and other organizational decision makers seek suggestions to improve their brand or organization~\cite{jijkoun-mining-2010}. 
In this case, consumers or other stakeholders are explicitly asked to provide suggestions. 
Opinions towards persons, brands, social debates etc. are generally expressed through online reviews, blogs, discussion forums, or social media platforms, and tend to contain the expressions of advice, tips, warnings, recommendations etc.~\cite{amigo-overview-2014}. 
For example, online reviews may contain suggestions for improvements in the product or service (Table~\ref{examples}); and recommendation platforms often ask for specific tips from their users, which are then offered to other users; see Figure~\ref{roomtips} for an example from travel site TripAdvisor.\footnote{\url{https://www.tripadvisor.com}}

\begin{table*}[h]
\caption{Examples of suggestions from different domains.}
\label{examples}
\centering
\begin{tabular}{>{\raggedright}p{2.2cm}p{8.8cm}}
\toprule
{\bf Source} & {\bf Sentence}\\
\midrule
Electronics reviews & I would recommend doing the upgrade to be sure you have the best chance at trouble free operation.\\
\midrule
Electronics reviews & My one recommendation to creative is to get some marketing people to work on the names of these things \\
\midrule
Hotel reviews & Be sure to specify a room at the back of the hotel.\\
\midrule
Twitter & Dear Microsoft, release a new zune with your wp7 launch on the 11th. It would be smart \\\hline
Travel discussion forum & If you do book your own airfare, be sure you don't have problems if Insight has to cancel the tour or reschedule it \\
\bottomrule
\end{tabular}
\end{table*}

State-of-the-art opinion mining systems primarily summarize these opinions as a distribution of positive and negative sentiments by means of sentiment analysis methods~\cite{liu2012sentiment}, and therefore suggestion mining remains a relatively young area. 
So far, it has usually been defined as a problem of classifying sentences of a given text into \emph{suggestion} and \emph{non-suggestion} classes. 
Suggestion mining faces similar challenges as other newly introduced sentence classification tasks. 
These include:
\begin{inparaenum}[(1)]
\item task formalization and data annotation, 
\item understanding sentence level semantics, 
\item figurative expressions, 
\item long and complex sentences, 
\item context dependency, and
\item highly imbalanced class distribution in some domains.
\end{inparaenum} 
As we will see below, the domains covered previously include hotel reviews, electronics reviews, Twitter, and travel discussion forums, with the majority of studies having focused on collecting suggestions for product improvement using product reviews as a source text. 
Problem definition and methods remain tailored for their specific application. 
Mostly rule-based systems have so far been developed, and very few statistical classifiers have been proposed.

In this study, we address the following research questions:
\begin{description}
\item[RQ1] How do we define suggestions in the context of open domain suggestion mining?
\item[RQ2] How do we prepare benchmark datasets for suggestion mining?
\end{description}

\noindent%
The main contributions of this paper are as follows:
\begin{inparaenum}[(1)]
\item Study of layman perception of the term \emph{suggestion};
\item Proposal for an empirically driven definition of suggestions;
\item Study of the linguistic properties observed in sentences labeled as suggestion as per our definition;
\item Proposition of an annotation method that employs both layman and expert annotators, with the aim of minimizing annotation time and cost without compromising the quality of datasets; and 
\item Development and distribution of benchmark datasets.
\end{inparaenum}

In Section~\ref{section:2} we discuss related work on problem definition and datasets for suggestion mining. 
Section~\ref{section:3} proposes a problem definition based on an annotation study performed with layman annotators. 
In Section~\ref{section:4} we propose a method to create benchmark datasets for suggestion mining,  inspired by our observations and proposed problem definition.
Section~\ref{section:5} concludes the paper.


\section{Related Work}\label{RelatedWork}
\label{section:2}

A common problem definition followed in different publications on suggestion mining is
\begin{quote}
\emph{Given a sentence s, predict a label l for s where l $\in$ $\{$suggestion, non suggestion$\}$}. 
\end{quote}
In this study, we aim to extend this problem definition by formally defining the scope of suggestion and non-suggestion classes in a manner that will suit both open domain and domain specific suggestion mining. 

Previous work that attempted to define suggestions, did so in two ways. 
One was to provide a dictionary-like generic definition \cite{Wicaksono2012,Viswanathan2011} such as \emph{A sentence made by a person, usually as a suggestion or a guide to action and/or conduct relayed in a particular context}. 
The other was to provide an application specific definition of suggestions \cite{Ramanand2010,Brun2013,Moghaddam2015} such as \emph{Sentences where the commenter wishes for a change in an existing product or service}. 
Although the first category is generic and applies to all domains, the publications listed evaluated suggestion mining on single domains. 
In our annotation study on several domains, on which we elaborate in the next section, we observe that a generic definition of suggestion leads to higher disagreement among the annotators. 
On the other hand, when a domain and use case specific definition was provided in related work, the formal annotation guidelines were still missing. 
Importantly, such definitions cannot be used to define the scope of open domain suggestion mining.

\citet{Wicaksono2012,Wicaksono2013a,Wicaksono2013b} performed the extraction of what they refer to as the \emph{advice revealing sentences} from travel related weblogs and online forums, using supervised learning methods. 
Their dataset is available to the research community.\footnote{Available upon request from the authors}
However, no formal annotation guidelines were provided \cite{Wicaksono2013b} during the annotation. The kappa statistics for inter-annotator agreement was 0.76, and only those sentences were included in the datasets on which the two annotators agreed. 
In order to prepare the dataset, they first filtered out a sample of blog entries by using clue words like \emph{suggest}, \emph{advice}, \emph{recommend}, \emph{tip}, etc. 
This approach of filtering the real data may bias the results towards the proposed features since the authors also used these clue words as features. 
Another publication that employed supervised learning was \citet{Dong2013}, who performed suggestion detection on tweets about the Microsoft Windows 7 phone. 
They did not define suggestions, but mentioned that the objective of collecting suggestions is to improve and enrich the quality and functionality of products, services, and organizations. 
The dataset prepared by them is also publicly available,\footnote{http://homepages.inf.ed.ac.uk/s1478528/} however the formal annotation details are again not available. 
The authors did not provide inter-annotator agreement scores, and retained only those tweets in the dataset where all the annotators agreed on the label. 

Datasets from other publications are not available for further investigation, mainly because of the proprietary nature of the data; see Table~\ref{existingDatasets}. 
\citet{Moghaddam2015} performed suggestion mining at the review level, by identifying suggestion containing reviews among a large number of reviews about the eBay App. 
They report that 5~human annotators were employed, and a 96\% agreement was observed in their annotations, where only agreed upon instances were retained in the dataset. 

None of the studies that proposed and evaluated rule-based systems for this task \cite{Ramanand2010,Viswanathan2011,Brun2013}, performed an annotation study and provided inter-annotator agreement scores. 

\begin{table*}[t]
\caption{An overview of related work and available datasets.}
\label{existingDatasets} 
\centering
\begin{tabular}{lll}
\toprule
\bf Publication & \bf Source & \bf Dataset available  \\ 
\midrule
\citet{Ramanand2010}& Product reviews & No \\
\citet{Viswanathan2011}& Product reviews & No \\
\citet{Brun2013}& Product reviews & No \\
\citet{Moghaddam2015}& Hotel reviews & No\\
\citet{Wicaksono2012,Wicaksono2013a,Wicaksono2013b}& Travel forum & Yes \\
\citet{Dong2013}& Twitter & Yes \\
\bottomrule
\end{tabular}
\end{table*}

The publications on suggestion mining listed above and summarized in Table~\ref{existingDatasets} did not investigate the definition and scope of suggestions as a research question. 
They did remove the disagreed instances from the train and test datasets if manual labeling was performed using the broad definitions as guidelines. 
Although, this may be one way to compensate for a missing annotation study, models trained on such unambiguous datasets may result in a lowered performance on real life data, as compared to the unambiguous test datasets. 
In this work, we perform a study of different perceptions of the term \emph{suggestion}, formalize the definition of suggestions, and propose a \emph{two phase} annotation method in order to create benchmark datasets.


\section{Problem Definition}
\label{section:3}

Existing problem definitions of \emph{suggestion mining} define the task as predicting \emph{suggestion} and \emph{non-suggestion} labels for sentences. 
While this definition applies to all domains and applications of suggestion mining, the scope of the \emph{suggestion} and \emph{non-suggestion} classes should also be defined. 

Many linguistic studies have investigated suggestions from an open domain, syntactic, pragmatic, or philosophical perspective. 
For example, according to \citet{Searle1976}, suggestions \emph{belong to the group of directive speech acts, which are those in which the speaker's purpose is to get the hearer to commit him/herself to some future course of action}. 
These studies relate to the classical and standard form of language, and so are the definitions of suggestions. We observe that such definitions are less suitable for suggestion mining. 
Firstly, in the context of text mining we are dealing with informal text on the web, which often involves a non-standard use of language. 
Secondly, linguists or expert annotators would be required to annotate datasets as per the definitions in linguistic studies. 
Thirdly, suggestion mining systems are highly likely to be used by layman users and therefore the predicted suggestions should be aligned with a layman understanding.

\citet{Wicaksono2013a} mentioned a broad and domain independent definition of suggestion as a `sentence that contains a suggestion for or guide to an action to be taken in a particular context.'  
They simply asked annotators to label \emph{advice-revealing} sentences in a given dataset. 
Our annotation study showed that human perception of the term \emph{suggestion} is subjective, and this effects the preparation of hand labeled suggestion datasets. 
We first present a qualitative and quantitative analysis of our annotation study below, and based on our observations propose a formal definition of both suggestion and non-suggestion classes, as well as an annotation approach for the benchmark datasets. 
In previous work~\cite{NegiEmnlp2015}, we proposed a formal task definition, but it was specific to the use case of mining customer to customer suggestions from online reviews.

\subsection{Annotation Study}
We study the perception of layman users towards the term \emph{suggestion} by performing a trial annotation where no formal annotation guidelines were provided, and the annotators were simply asked to choose between the labels \emph{suggestion} and \emph{non-suggestion} for a given sentence. 
We collected a total of 80~sentences, 20~sentences each from 4~domains: hotel reviews, restaurant reviews, software developers' suggestion forum, and tweets. 
We made sure that at least 5 potential suggestion sentences were present in each of these four sets. 
In the case of tweets, the entire tweet is considered as a single instance instead of a sentence. 
Each sentence was annotated by 50 layman annotators recruited through Crowdflower.\footnote{Known as Crowdflower (\url{https://www.crowdflower.com}) during the time of study, the platform is re-branded as Figure8 now (\url{https://www.figure-eight.com/}) } 
We also provided the context along with these sentences, where context refers to the source of the sentence, as well as the text of the entire document to which a sentence belongs. 
For example, context comprises of the entire review text in the case of reviews, and the entire post in the case of suggestion forum. 

\begin{table*}[t]
\caption{Examples of sentences accepted as suggestions by layman annotators. $\%$ Annotators in each table refers to the percentage of total annotators who labeled a given sentence as a suggestion.}
\label{SuggExample} 
\begin{tabular}{p{5.6cm}>{\raggedright}p{1.6cm}>{\raggedleft}p{1.3cm}p{1.7cm}}
\toprule
\bf Sentence & \bf Domain &\bf $\%$ Annotators & \bf Category  \\ 
\midrule
We did not have breakfast at the hotel but opted to grab breads/pastries/coffee from the many food-outlets in the main stations.& Hotel review & 80 &Tell \\
\midrule
I recommend going for a Trabi Safari & Hotel review & 100 & Recommend \\
\midrule
Definitely come with a group and order a few plates to share.& Restaurant review & 52 & Suggest \\
\midrule
Please provide consistency throughout the entire Microsoft development ecosystem!" & Suggestion forum & 70 & Request \\
\midrule
We need a supplementary fix to the issues faced by existing affected users without resorting them to contacting Microsoft customer support & Suggestion forum & 60 & Need/ Necessity\\
\midrule
RT \@larrycaring: "every time someone send you hate send them this back" this is v good advice I'll use it every time antis insult me & Twitter & 84 & Suggest, Appreciation \\
\midrule
There is a parking garage on the corner of Forbes so its pretty convenient.& Restaurant review & 62 & Inform, Appreciation \\
\bottomrule
\end{tabular}
\end{table*}

Tables~\ref{SuggExample} and \ref{NonSuggExample} provide examples of the sentences that were perceived as suggestions and non-suggestions, respectively, by more than 50\% of the annotators. 
Unlike previous studies, we also study the characteristics of sentences labeled as non-suggestions. 
In order to define the scope of sentences considered as suggestions and non-suggestions by these annotators, we match the labeled sentences with their potential opinion expression category following the categorization defined by \citet{Asher2009}. 
These categories are \emph{Inform}, \emph{Assert}, \emph{Tell}, \emph{Remark}, \emph{Think}, \emph{Guess}, \emph{Blame}, \emph{Praise}, \emph{Appreciation}, \emph{Recommend}, \emph{Suggest}, \emph{Hope}, \emph{Anger}, \emph{Astonishment}, \emph{Love}, \emph{Hate}, \emph{Fear}, \emph{Offence}, \emph{Sadness/Joy}, \emph{Bore/Entertain}. 
A sentence with multiple clauses may contain more than one expression. 
We observe that most of the suggestions fell under the category of \emph{Suggestion}, \emph{Recommendation}, \emph{Request}, and \emph{Need/Necessity} opinion expressions. 
Annotators tend to choose \emph{Suggestion} if they inferred a suggestion from the sentence, even if the sentence did not contain the expressions typically used in a suggestion. 
For example, some sentences with `Tell' expressions were labeled as suggestions (Table~\ref{SuggExample}), which is a dominant category of expressions in non-suggestions (Table \ref{NonSuggExample}). 
Also, there was a high disagreement over many sentences ($>$30\%)  

\begin{table*}[t]
\caption{Examples of sentences accepted as non-suggestions by layman annotators.}
\label{NonSuggExample} 
\begin{tabular}{p{5.4cm}p{1.8cm}>{\centering}p{1.6cm}p{1.4cm}}
\toprule
\bf Sentence & \bf Domain & \bf $\%$ Annotators & \bf Category \\ 
\midrule
This is much safer and far more convenient than hailing taxis from the street. & Hotel Review & 70 & Assert \\
\midrule
There is a very annoying bug in the Windows 10 store that hides apps from listing.& Suggestion Forum & 64 & Tell \\
\midrule
Just returned from a 3 night stay.& Hotel Review & 94 & Tell \\
\midrule
The internet was free but I do think people should only be allowed say 30 mins and its written in a book, thats a fair way I think.& Hotel Review & 72 & Tell, Think, Suggest \\
\midrule
I had so much fun filming all kinds of tips and advice (and some yoga!) yesterday & Twitter & 84 & Inform, Tell, Entertain \\
\bottomrule
\end{tabular}
\end{table*}

In order to define the scope of suggestions and formulate annotation guidelines for benchmark datasets, one approach is to identify which of the 20 expressions types by by \citet{Asher2009} map to the suggestion and non-suggestion classes. 
However, identifying suggestions on the basis of the type of opinion expression in a sentence may require a high level of linguistic understanding by the annotators. 
Also, certain type of expressions are present in both suggestions and non-suggestions.

We then provided some guidelines in a second round of annotation study. 
The guidelines asked the annotators to only label sentences where suggestions were explicitly expressed and not inferred. 
Examples of explicit and implicit expression of sentences were also provided. 
No improvement in disagreement was observed when detailed guidelines were provided. 
We observed that workers on crowdsourcing platforms tend to ignore or not comprehend the detailed guidelines. 
On the one hand, crowdsourcing with layman annotators may not be the best suited method for creating suggestion mining benchmark datasets, while on the other hand, trained annotators are either slower (volunteers) or expensive.

Apart from the detailed guidelines, we observed that context plays some role in the judgment of the annotators, since many annotators decided the labels based on the source text and not solely on the given sentence. 
Therefore, we also performed a round of annotations without providing the context. 
The following trends were observed:
\begin{itemize}
\item There was a tendency to label sentiment sentences such as \emph{They serve really nice breakfast} as suggestions when the context was not provided. 
\item When the context was provided, and a large number of sentences in the source text were expressing sentiments, annotators tend to choose more explicitly expressed sentences as suggestions. 
\item In cases where the sentence did not contain any information about the reviewed entity, implicit suggestions were marked as non-suggestions. For example, for \emph{There is a parking garage on the corner of Forbes so its pretty convenient} the agreed upon label changed to a non-suggestion when the restaurant review context was not provided.
\end{itemize} 
Based on these observations, we propose an extended problem definition that defines the scope of suggestion and non-suggestion sentences.

\subsection{Task Definition Revisited}
As we saw in our layman annotation study, context may affect an annotator's judgment. 
In the absence of context, different annotators associate different contexts to a candidate sentence. 
We observe that the following concepts form an integral part of defining a suggestion for suggestion mining.
\begin{description}
    \item[\em Surface structure.] 
    Different surface structures~\cite{Chomsky1957,Crystal2011} can be used to express the underlying intention of giving the same suggestion. For example, \emph{The nearby food outlets serve fresh local breakfast and are also cheaper} and \emph{You can also have breakfast at the nearby food outlets which are cheaper and equally good}.
    Linguistic studies in the past studied the way suggestions are expressed (see Table~\ref{MartinezSuggestionTaxonomy}) and provided the examples of lexical elements used in the surface structures for suggestions. 
    \item[\em Context.] When dealing with specific use cases, context plays an important role in distinguishing a suggestion from a non-suggestion. Context may be present within a given sentence. It can be a set of values corresponding to different variables that are provided explicitly and in addition to a given sentence. 
    One or more of the following variables can constitute the context:
    \begin{description}
        \item[\em Domain.] We refer to the source of a text as \emph{domain}. In the process of dataset annotation, we closely studied some of the domains; Table~\ref{datasets} shows how the distribution of suggestions varies with the domains. 
        \item[\em Source text.] The text in the entire source document to which a sentence belongs may also serve as a context, giving an insight into the discourse where the suggestion appeared.
        \item[\em Application or use case.] Suggestions may sometimes be sought only around a specific topic, for example, mining room tips from hotel reviews. Suggestions can also be selectively mined for a certain class of users, for example, mining suggestions for future customers. 
        All non-relevant suggestions in the data may be regarded as non-suggestions in this case. 
        Previous studies on suggestion mine from online reviews operated on this kind of context. 
        For example, only suggestions for improvement were identified from the product reviews, whereas suggestions meant for the fellow customers \cite{NegiEmnlp2015} were considered as non-suggestions.
    \end{description}
\end{description}
We now propose an empirically driven and context-based definition of suggestions.
Given that
\begin{itemize}
\item $s$ denotes the surface structure of a sentence,
\item $C$ denotes additional context provided with $s$, where the context can be a set of values corresponding to certain variables, and
\item $a(s,C)$ denotes the annotation agreement for the sentence, and $t$ denotes a threshold value for the annotation agreement,
\end{itemize}
we write $S(s,C)$ to denote the \emph{suggestion function}, which is defined as
\begin{equation} \label{eq4_1}
S(s,C) = \begin{cases}
Suggestion,\, \text{ if } a(s,C) \geq t \\
Non\text{-}suggestion, \text{ if }a(s,C) < t .
\end{cases}
\end{equation}

\noindent%
Depending on the choice of $C$, and, hence, on the value of $a(s,C)$, we identify four categories of sentences that a suggestion mining system is likely to encounter. 

\begin{description}
\item[\em Explicit suggestions.] \emph{Explicit suggestions} are sentences for which $S$ always outputs \emph{Suggestion}, whether $C$ is the empty set or not.
The suggest, recommend, request, need/necessity category of opinion expressions in Table~\ref{SuggExample} are mostly found in the explicit expressions of suggestions. 
They are like the \emph{direct} and \emph{conventionalised} forms of suggestions as defined by \citet{Martinez2005} (Table~\ref{MartinezSuggestionTaxonomy}). 
It may also be the case that such sentences have a strong presence of context within their surface form, as in illustrated by \emph{If you do end up here, be sure to specify a room at the back of the hotel}.

\item[\em Explicit non-suggestions.] These are the sentences for which $S$ always outputs \emph{Non-suggestion}, whether $C$ is the empty set or not. For example, \emph{Just returned from a 3 night stay}.

\item[\em Implicit suggestions.] These are sentences for which $S$ outputs \emph{Non-suggestion} only when $C$ is the empty set.
Typically, implicit suggestions do not posses the surface form of suggestions but the additional context helps the readers to identify them as suggestions.
For example, \emph{There is a parking garage on the corner of Forbes, so its pretty convenient} is labeled as a suggestion by the annotators when the context is revealed as that of a restaurant review. 
A sentence such as \emph{Malahide is a pleasant village-turned-dormitory-town near the airport} can be considered as a suggestion given that it is obtained from a travel discussion thread for Dublin. 
These kind of sentences are observed to have a lower inter annotator agreement than the above two categories.

\item[\em Implicit non-suggestions.] These are sentences for which $S$ outputs \emph{Suggestion} only when $C$ is an empty set. 
Typically, an implicit non-suggestion posses the surface form of suggestions but the context leads readers to identify them as non-suggestions. 
Such sentences may contain sarcasm. Examples include \emph{Do not advertise if you don't know how to cook} appearing in a restaurant review and \emph{The iPod is a very easy to use MP3 player, and if you can't figure this out, you shouldn't even own one} appearing in a MP3 player review.
\end{description}

\noindent%
Based on the above four categories, we can define the scope of the suggestion and non-suggestion classes for open domain suggestion mining. 
For open domain suggestion mining, we propose to limit the scope of suggestions to the \emph{explicit suggestions}. Therefore, we set the task definition for \emph{open domain} suggestion mining to be:

\begin{itemize}
\item[] Let $s$ be a sentence. If $s$ is an explicit suggestion, assign the label \emph{Suggestion}. Otherwise, assign the label \emph{Non-suggestion}.
\end{itemize} 

\noindent%
The proposed categories provide the flexibility to change the scope of classes in a well defined manner, as well as to define context as per the application scenario. 

\subsection{Linguistic Observations}
Explicit suggestions are often expressed by means of certain lexical cues and grammatical moods. 
They tend to contain certain keywords and phrases, like \emph{suggest}, \emph{suggestion}, \emph{recommendation}, \emph{advice}, \emph{I suggest}, \emph{I recommend}, etc. 
Most of the previous work created a hand crafted list of such words and phrases to use them as features with the classifiers. 
However, not all the suggestions contain these keywords and phrases. 
Table~\ref{rankedNgrams} lists the top 10 unigrams and bigrams in the sentences tagged as advice and suggestion in the dataset used by \citet{Wicaksono2013a} and \citet{Dong2013}, and some examples which exclude such keyphrases. 
The ranking is based on the frequency count and unigrams exclude the stopwords.

\begin{table*}[h]
\caption{Three types of expressions of suggestions as defined by Martinez (2005) \cite{Martinez2005} with the examples of prevalently used surface structures.}
\label{MartinezSuggestionTaxonomy}
\centering
\begin{tabular}{ll}
\toprule
{\bf Type} & {\bf Example} \\
\midrule
\multirow{6}{*}{Direct} & I suggest that you \ldots \\
& I recommend that you \ldots \\
& I advice you to \ldots  \\
& My suggestion would be \ldots \\
& Try using \ldots \\
& Don't try to \ldots \\\hline
\multirow{6}{*}{Conventionalised forms} & Why don't you \ldots ?\\
& How about \ldots ?\\
& Have you thought about \ldots ?\\
& You can \ldots \\
& You could \ldots \\
& If I were you, I would \ldots \\\hline
\multirow{6}{*}{Indirect}& One thing (that you can do) would be \ldots \\
&Here's one possibility \ldots \\
& There are a number of options that you \ldots \\
& It would be helpful if you\ldots \\
& It might be better to \ldots  \\
&It would be nice if \ldots \\
\bottomrule
\end{tabular}
\end{table*}
%
%
\begin{table*}[h]
\caption{Top 10 unigrams and bigrams in the sentences labeled as advice and suggestion in the dataset provided by \citet{Wicaksono2013a} and \citet{Dong2013}, respectively.}
\label{rankedNgrams}
\centering
\begin{tabular}{llll}
\toprule
\multicolumn{2}{c}{\bf Travel} & \multicolumn{2}{c}{\bf Microsoft Tweets} \\
\cmidrule(r){1-2}\cmidrule(l){3-4}
{\bf Unigrams} & {\bf Bigrams} & {\bf Unigrams} & {\bf Bigrams}\\
\midrule
You & credit card & Microsoft& Windows Phone\\
tour& you will &Windows&Dear Microsoft\\
if& if want &WP7&Microsoft needs\\
just& http www&phone& Come Microsoft \\
travel& make sure &need& Microsoft make\\
time&Europe board &nokia&Microsoft WP7 \\
like&tour director&make& If Microsoft\\
hotel&post Europe&needs&Microsoft really\\
did&travel tips&apps&really needs\\
need&United States&want&hope Microsoft\\
\bottomrule
\end{tabular}
\end{table*}

\paragraph{Mood.} 
Suggestion expressions often contain what may be referred to as \emph{subjunctive} and \emph{imperative moods}~\citep{Morante2012,NegiEmnlp2015}. 
Subjunctive mood is a commonly occurring language phenomenon in Indo-European languages, which is typically used in subordinate clauses to express an action that has not yet occurred, in the form of a wish, possibility, necessity etc. \cite{Guan2012}. 
Typical examples include \emph{If the Duke were here he would be furious}~\cite{Dudman1988} and \emph{It is my suggestion that the students be sent to Tibet}~\citep{Guan2012}. 
The imperative mood is used to express the requirement that someone perform or refrain from an action. 
A typical example would be \emph{Take an umbrella}, \emph{Pray everyday}~\cite{Portner2009}. \citet{Wicaksono2012,Wicaksono2013a,Wicaksono2013b} and \citet{Dong2013} used the imperative mood as a feature with their classifiers. 
However, subjunctive mood has not been associated with suggestions in previous suggestion mining studies. 
 
\paragraph{Sentiment.} 
In the context of opinion mining, suggestion mining and sentiment analysis can be applied to very similar domains or data sources, for example, reviews, blogs, discussion forums, twitter etc. 
Sentiment expressions and suggestion expressions may also appear together in the same sentence, especially in the case of recommendation type of suggestions. 
In sentiment analysis, text is generally categorized into three or more classes, while in suggestion mining a text is either a suggestion, or a non-suggestion. 
Suggestion expressing texts are not limited to a particular class of sentiment. 
While sentiment sentences are always defined as subjective, suggestions can be found in both objective and subjective type of text. Therefore, a suggestion bearing sentence may be associated to multiple sentiments. 

\begin{figure}[h]
\centering
\includegraphics[scale=0.65]{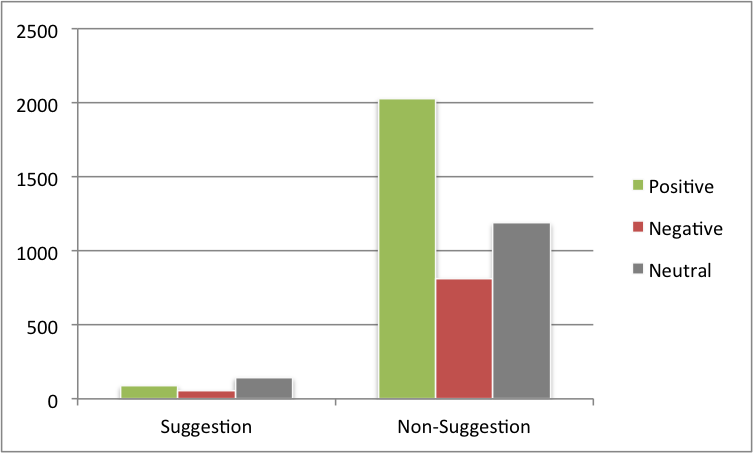}
\caption{Distribution of sentiment polarities towards the hotel in suggestion and non-suggestion sentences for a hotel review dataset~\citep{Wachsmuth2014}.}
\label{SuggAndSentHotel}
\end{figure}

In the case of reviews, some sentiment analysis benchmark datasets like SemEval datasets \cite{Pontiki2015} exclude text that is not about the opinion target, even though the text is found within the same review. 
The guidelines from the SemEval 2015 Sentiment Analysis task \cite{Pontiki2015} read: \emph{Quite often reviews contain opinions towards entities that are not directly related to the entity being reviewed, for example, restaurants/hotels that the reviewer has visited in the past, other laptops or products (and their components) of the same or a competitive brand. Such entities as well as comparative opinions are considered to be out of the scope of SE-ABSA 15. In these cases, no opinion annotations were provided.} 
Some of the non-relevant sentences in review datasets are potential suggestions on closely related points of interest. 
Suggestions in hotel reviews may contain tips and advice about the reviewed entity, suggestions and recommendations about the neighborhood, transportation, and things to do. 
Similarly, in product reviews, suggestions can be about how to make a better use of the product, accessories which go with them, or availability of better deals, etc. (Table~\ref{SuggestionAndSentiments}). 
Figure~\ref{SuggAndSentHotel} shows the distribution of suggestions and sentiments (expressed towards the hotel) in a hotel review dataset annotated with sentiments by \citet{Wachsmuth2014}. \\

\begin{table}
\caption{Examples of suggestions from review datasets for sentiment analysis. \emph{Sentiment} refers to the sentiment towards the reviewed entity, while \emph{Relevant} indicates if the suggestion is relevant for calculating the sentiment towards the reviewed entity.}
\label{SuggestionAndSentiments} 
\centering
\begin{tabular}{p{5.8cm}p{1.5cm}p{1.4cm}p{1.4cm}}
\toprule
\bf Sentence & \bf Domain & \bf Sentiment & \bf Relevant \\ 
\midrule
One more thing- if you want to visit the Bundestag it is a good idea to book a tour (in English) in advance & Hotel & Neutral & No\\
\midrule
Be sure to pick up an umbrella (for free) at the concierge if you anticipate rain while sightseeing.& Hotel & Positive & Yes\\
\midrule
You are going to need to buy new headphones, the stock ones suck & MP3 player & Negative & Yes \\
\midrule
If you strictly use the lcd and not the view finder , i highly recommend the camera .& Camera & Positive & Yes \\
\midrule
For those of you who already bought this camera , I suggest you buy a hi-ti dye-sub photo printer
& Camera & Neutral & No\\
\midrule
If only it played stand alone avi files & DVD player & Neutral & Yes \\
\midrule
Would be really good if they have given an option to stop this auto-focusing & Camera & Neutral & Yes\\\hline
\end{tabular}
\end{table}

\noindent%
In this section we answered RQ1, i.e., \emph{How do we define suggestions in the context of open domain suggestion mining?}
We provided a formal definition for suggestions by identifying four categories of sentences, i.e., explicit suggestions, implicit suggestions, explicit non-suggestions and implicit non-suggestions, where the explicit suggestions are defined as suggestions with regards to open domain suggestion mining. 


\section{Creating Benchmark Datasets for Suggestion Mining}
\label{section:4}
\label{GoldDatasets}
\label{Data}

Based on our preliminary annotation study, problem definition and scope of suggestions, we propose a two phase method for manually annotating sentences with the class labels:
\begin{description}
\item[\em Phase 1] This phase is performed using paid crowdsourcing, where each sentence is annotated by multiple layman annotators.
\item[\em Phase 2] This phase is performed by an expert annotator. 
\end{description}
Below, we describe both phases as well as the subsequent creation of multiple datasets for suggestion mining.

\subsection{Phase 1: Crowdsourced Annotations}

We used Crowdflower\footnote{The company was re-branded as Figure Eight post this study. \url{https://www.figure-eight.com/}} to collect layman annotations.

\begin{figure}
\centering
\includegraphics[width=\textwidth]{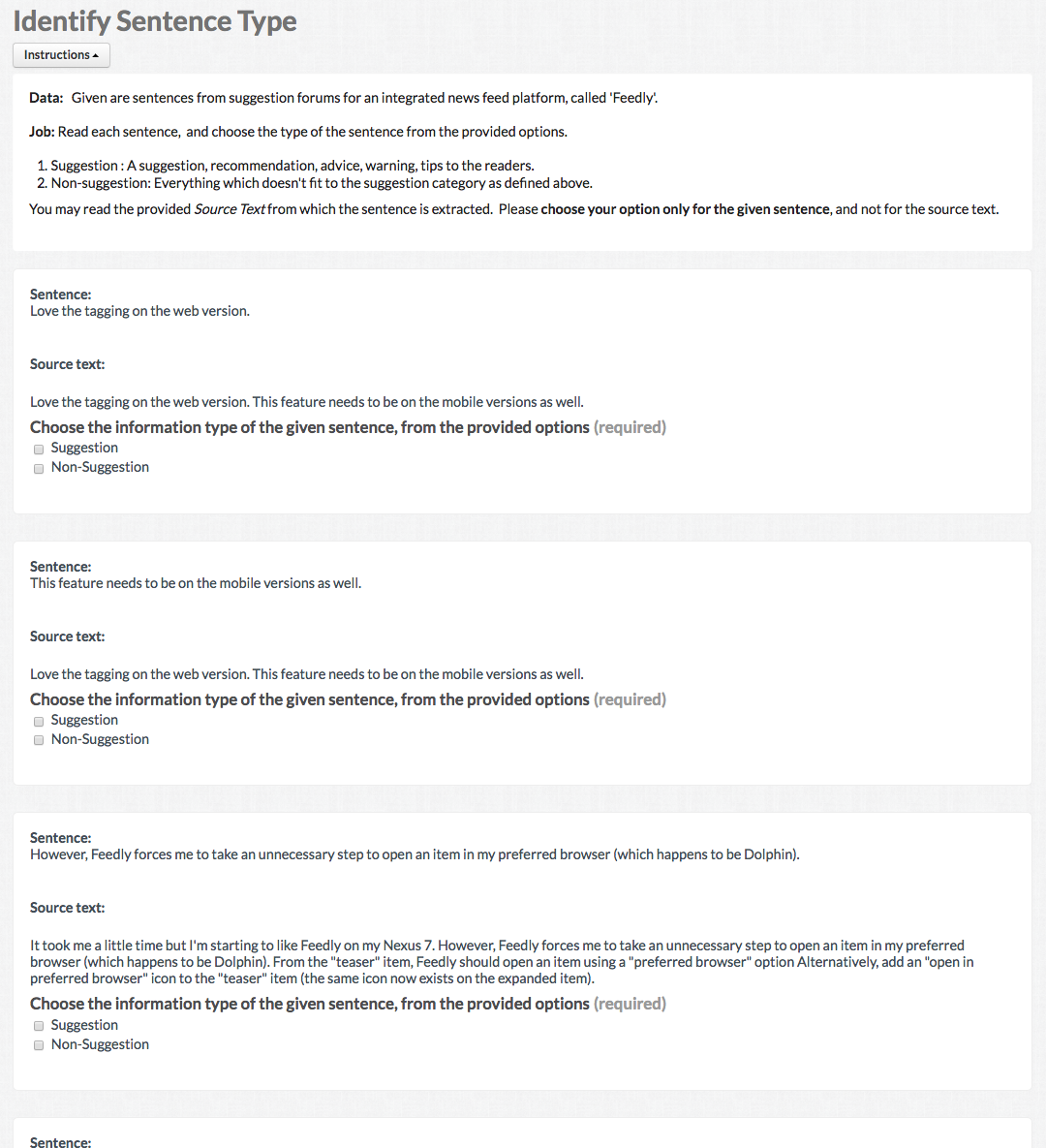} 
\caption{A screenshot of how each page of the annotation job appears to the annotators on the crowdflower platform.}
\label{crowdflowerJob}
\end{figure}

\paragraph{Job design.}
The annotation job was offered to annotators in batches of rows which are referred to as `Pages' (see Figure \ref{crowdflowerJob}). 
Annotators were not paid per individual row but per page. In our case, each page had 8 sentences. 
For quality control, before being allowed to perform a job, the annotators were presented with a set of test sentences which are similar to the actual questions except that their answers have already been provided by us to the system. 
We also submitted the explanation behind the correct answer. 
This way the test questions serve two purposes: test the annotators competency and understanding of the job, and train the annotator for the job. 

We submitted 30 test questions for each dataset. 
Each starting annotator was presented with 10 test questions, and only the annotators who score 70\% or more were allowed to proceed with the job. 
If an annotator passed the test and started the job, the remaining unseen test questions were presented to them in between the regular sentences, and without being notified. 
One question out of 8~on a page was a hidden test question for the annotator. 
The accuracy score of a contributor on test questions is referred to as \emph{Trust score} in a job. 
If an annotator's trust score dropped below a certain threshold during the course of the annotation, the system did not allow them to proceed further with the job. 
This threshold score in our case was set to 70\%. 

In addition to the hidden test questions, a minimum time for each annotator to stay on one page of the job was set.  
We set this time to 40 seconds (5 seconds on average for each sentence) for our jobs.
If annotators appeared to be faster than that, they were automatically removed from the job.

We restricted access to annotators from countries where English is a popular language and that are also likely to have a large crowdsourcing workforce. 
Most of the annotators came from Australia, Canada, Germany, India, Ireland, the United Kingdom, and the USA.

Crowdflower assigns workers with experience levels based on the number of times they successfully passed test questions and performed a job. 
We only allowed workers with experience level 2 or more for our annotations. 
Therefore, the annotation reward was required to be competent with other jobs. 
We paid 10~cents for each page, i.e., 8~sentences, which means a maximum of 10 cents for 40 seconds of work, which amounts to \$9 per hour\textcolor{red}, which respected the wage regulations of the country in which the first author resided at the time of writing.

\paragraph{Annotation agreement.}
We used Crowdflower to collect multiple judgments per sentence. 
Using Crowdflower's \emph{confidence} score, which describes the level of agreement between multiple contributors and the confidence in the validity of the result at the same time, we used a threshold confidence score of 0.6.
However, it can be the case that a sentence is very ambiguous and cannot achieve the confidence score even after a large number of workers answered it. 
A maximum limit to the number of annotators is set in such case, and no further judgements are collected even if the threshold confidence is not reached. 
We set this limit to 5 annotators. 
Sentences that do not pass the confidence threshold of 0.6 are counted as non-suggestions in the final dataset.

\subsection{Phase 2: Expert Annotations}
As mentioned previously, we follow a two phase annotation strategy, where one phase (Phase~1) includes the context and the other (Phase~2) excludes the context of a sentence. 
Our scope of suggestions is limited to sentences that are labeled as suggestion in both the phases, i.e., explicit suggestions.

In the crowdsourced annotations (Phase~1), the context was provided to the annotators in the form of domain information and source text. 
Phase~2 of the annotation is only applied to sentences that were labeled as suggestions in Phase~1. 
This drastically reduces the number of annotations to be performed in Phase~2. 
The inter-annotator agreement for Phase~2 was calculated by having two annotators label a subset of sentences for each domain (50~sentences). 
Cohen's kappa coefficient was used to measure the inter-annotator agreement. 
The remainder of the data instances were annotated by only one annotator. 

The following guidelines were provided to the annotators in Phase~2 :
\begin{itemize}
\item The intent of giving a suggestion and the suggested action or recommended entity should be explicitly stated in the sentence. 
\emph{Try the cup cakes at the bakery next door} is a positive example.
Other explicit forms of this suggestion could be: \emph{I recommend the cup cakes at the bakery next door} or \emph{You should definitely taste the cup cakes from the bakery next door}. 
An implicit way of expressing the suggestion could be \emph{The cup cakes from the bakery
next door were delicious}.
\item The suggestion should have the intent of benefiting a stakeholder and should not be
mere sarcasm or a joke. 
For example, \emph{If the player doesn't work now, you can run it over with your car} would not pass this test.
\end{itemize}

\noindent%
Following are some of the scenarios of conflicting judgments observed in this phase of annotation:
\begin{itemize}
    \item In the case of suggestion forums for specific domains, like a software developer forum, domain knowledge is required to distinguish an implicit non-suggestion from an explicit suggestion. 
    Consider, for example, the two sentences, \emph{It needs to be an integrated part of the phones functionality, that is why I put it in Framework} and \emph{Secondly, you need to limit the number of apps that a publisher can submit with a particular key word}. 
    The first sentence is a description of already existing functionality and is a context sentence in the original post, while the second is suggestion for a new feature.
    \item No concrete mention of what is being advised such as in \emph{It'd be great if you would work on a solution to improve the situation}. 
    \item A sentence such as \emph{I would go in fall} was annotated as a suggestion in Phase~1, as a part of travel discussion forum, with a likely interpretation as \emph{If I were you, I would go in the fall}. However, when viewed without context, it can be perceived as reporting one's travel plans.
    \item At times, there was a confusion between information (fact) or suggestion (opinion). For example, \emph{You can get a ticket that covers 6 of the National Gallery sites for only about US\$10}.
\end{itemize}

\noindent%
In the final versions of the datasets prepared by us (see below), the sentences that are labeled as suggestions in Phase~2 of the annotation process are labeled as suggestions, while all other sentences are labeled as non-suggestions. 

\subsection{New Datasets for Suggestion Mining}
This section lists the manually labeled datasets that were created following Phase~1 and Phase~2 described above. 

We apply our two-stage annotation process to create four new datasets for suggestion mining. On top of that we re-use existing datasets, viewing the labels originally provided as Phase~1 annotations of our two-stage annotation process.

\begin{description}
\item[\em Hotel reviews.] \citet{Wachsmuth2014} provide a large
dataset of hotel reviews from the TripAdvisor\footnote{\url{https://www.tripadvisor.com/}} website. 
They segmented the reviews into statements so that each statement has only one sentiment label and have manually labeled the sentiments. 
Statements are equivalent to sentences, and comprise of one or more clauses. 
These statements have been manually tagged with \emph{positive, negative, conflict, and neutral} sentiments. 
We take a smaller subset of these reviews, where each statement is considered as a sentence in our dataset. 

\item[\em Electronics reviews.] \citet{Hu2004} provide a dataset comprising of reviews of different kinds of electronic products obtained from the website of Amazon.\footnote{\url{https://www.amazon.com/}} The Amazon website collects and displays online reviews of listed products. 
\citeauthor{Hu2004} split the reviews into sentences; sentiment for each sentence has been manually tagged.

\item[\em Travel forum.] The data is obtained from a previous travel forum dataset by \citet{Wicaksono2013a,Wicaksono2013b}. This domain exhibits a wide variety of expressions employed in suggestions, with relatively lower grammatical and spell errors. 
However, this domain also shows a relatively lower inter-annotator agreement. 

\item[\em Software suggestion forum.] The sentences for this dataset were scraped from the Uservoice\footnote{\url{https://www.uservoice.com/}} platform. Uservoice provides customer
engagement tools to brands, and therefore hosts dedicated suggestion forums for certain products. 
The Feedly mobile application forum and the Windows developer forum are openly accessible. 
A sample of posts were scraped and split into sentences using the Stanford CoreNLP toolkit \cite{Klein2003}. 
The class ratio in the dataset obtained from suggestion forums is more balanced than the other domains. Many suggestions are in the form of requests, which is less frequent in other domains. 
The text contains highly technical vocabulary related to the software which is being discussed, which may effect the classifier performance when this dataset is used for training or evaluation in the cross domain train-test settings, specially when bag of word features are employed. 

\end{description}

\noindent%
Following are the reasons for choosing these four domains for datasets. Online reviews is a popular target domain for opinion mining and a number of sentiment tagged review datasets are available from previous studies, therefore we also prepared a suggestion mining datasets for electronics (product) and hotel (service) reviews. This also allows us to study the relationship between suggestions and sentence level sentiments in online reviews. Also, product reviews were popularly used in the related work for suggestion mining. The choice of travel forum dataset was also inspired from the related work on advice mining \cite{Wicaksono2012}. Software suggestion forum was chosen because review datasets were highly imbalanced for explicit suggestions, while suggestion forums had a better presence of explicitly expressed suggestions. Also, suggestion forum datasets supported a different use case of suggestion mining than the review datasets, i.e. summarisation of suggestion posts.

In addition to these four new datasets, we re-labeled two existing datasets for suggestion mining. We consider the labels provided from in previous work as Phase~1 annotations, and performed Phase~2 annotations on these datasets, i.e., re-labeling of instances that were previously labeled as suggestions. 

\begin{description}
\item[\em Travel forum dataset.] \citet{Wicaksono2013a,Wicaksono2013b} crawled several web forum threads
from two well-known travel forums InsightVacations\footnote{\url{http://forums.insightvacations.com}} and Fodors.\footnote{\url{http://www.fodors.com}} 
Originally, an inter-annotator agreement of 0.76 (Cohen's kappa) was reported for this dataset. 
The provided dataset only comprises of the instances where both the annotators agreed.

\item[\em Microsoft tweets dataset.] The dataset was initially released by \citet{Dong2013}. While no inter-annotator agreement was reported by the authors, only those tweets were retained in the dataset where the annotators mutually agreed upon the label. 
In this case, the unit for suggestions is a tweet instead of a sentence.  
All of the tweets previously labeled as suggestions in the Microsoft tweet dataset were accepted as suggestions in Phase~2 annotations. 
A 100\% inter-annotator agreement was observed between the two annotators. 
This could be due to the fact that full tweet is available to the annotators rather than a single sentence, while hashtags also help in reducing ambiguities.
\end{description}

\begin{table}
\caption{Manually annotated suggestion mining datasets created in this paper.}
\label{datasets}
\centering
\begin{tabular}{p{2.8cm}p{3.5cm}>{\raggedleft}p{2.5cm}p{1.4cm}}
\toprule
\bf Dataset identifier & \bf Source & \bf Suggestion : Non-suggestion & \bf Inter-annotator agreement (Phase~2)\\
\midrule
\multicolumn{4}{c}{Existing datasets}\\
\midrule
\raggedright Microsoft tweets (original, re-tagged) & Twitter & 238/2762 (0.08) & 1.0 \\
\raggedright Travel forum (original) & InsightVacations, Fodors & 2192/3007 (0.72) & 0.76 \cite{Wicaksono2013a}\\
\raggedright Travel train (re-tagged) & InsightVacations, Fodors & 1314/3869 (0.34) & 0.72 \\
\midrule
\multicolumn{4}{c}{New datasets}\\
\midrule
Hotel train & Tripadvisor& 448/7086 (0.06) & 0.86 \\
Hotel test & Tripadvisor& 404/3000 (0.13) & 0.86 \\
Electronics train& Amazon & 324/3458 (0.09) & 0.83 \\
Electronics test& Amazon & 101/1070 (0.09) & 0.83 \\
Travel test & Fodors & 229/871 (0.26) & 0.72 \\
Software train & Uservoice suggestion forum & 1428/4296 (0.33) & 0.81 \\
Software test & Uservoice suggestion forum & 296/742 (0.39) & 0.81 \\
\bottomrule
\end{tabular}
\end{table}

\noindent%
Table~\ref{datasets} provides the details of the suggestion mining datasets created as part of this work. The datasets\footnote{\url{http://server1.nlp.insight-centre.org/sapnadatasets/ThesisDatasets/}} are freely made available for non-commercial purposes.\footnote{\url{https://creativecommons.org/licenses/by-nc-sa/4.0/}}

In this section we answered RQ 2, i.e., \emph{How do we prepare benchmark datasets for suggestion mining}. 
We studied the annotation challenges associated with suggestion mining, and proposed a two phase annotation method for preparing datasets for open domain suggestion mining. 
Phase~1 provides context with the sentences to be annotated and is performed by layman annotators using a crowdsourcing platform, while Phase~2 does not reveal the context to the annotators, and is performed by expert annotators. 
The sentences that were labeled as suggestions in both phases are retained as suggestions while the rest of the sentences are labeled as non-suggestions.


\section{Conclusion} 
\label{section:5}

In this paper we have focused on answering two main research questions, viz.\ how to define suggestions in the context of open domain suggestion mining, and how to create benchmark datasets for suggestion mining. 

To inform our annotation effort, we report on a study of the perception of the term \emph{suggestion} among layman annotators. 
We map the sentences labeled as suggestions and non-suggestions by layman annotators to some predefined categories of expressions which tend to appear in opinion discourse. 
These categories were thoroughly studied and defined in existing work \cite{Asher2009}. 
Some of the categories of expressions were present in both suggestion and non-suggestion sentences, which forms the basis of ambiguities associated with the preparation of benchmark datasets. 

Based on the observations, we propose a context dependent method to define four categories of sentences encountered in a source text, explicit suggestions, implicit suggestions, explicit non-suggestions, an implicit non-suggestions. We also study the possible contexts which can play a significant role in determining whether a sentence should be regarded as a suggestion or not. Based on this theory, we develop benchmark datasets for suggestion mining using a two-phase annotation method. A detailed account is shared on the methodology we adapted to prepare benchmark datasets for suggestion mining using a combination of crowdsourced and expert annotators. 

Finally, we release new benchmark datasets for suggestion mining relating to five domains, that are publicly available for research purposes.  
%
%
\bibliographystyle{spbasic}
\bibliography{suggestion-mining}

\end{document}